\documentclass{article}

\usepackage{microtype}
\usepackage{graphicx}
\usepackage{subfigure}
\usepackage{booktabs} 
\usepackage{multirow}
\usepackage{lmodern} 
\usepackage{microtype}
\usepackage[numbers,square,sort&compress]{natbib}
\usepackage{xurl}  
\usepackage{bm} 
\usepackage{hyperref}

\usepackage{hyperref}

\usepackage[accepted]{icml2023}

\usepackage{amsmath}
\usepackage{amssymb}
\usepackage{mathtools}
\usepackage{amsthm}
\usepackage{adjustbox}
\usepackage[english]{babel}

\usepackage[capitalize,noabbrev]{cleveref}

\theoremstyle{plain}

\theoremstyle{definition}

\theoremstyle{remark}

\usepackage[textsize=tiny]{todonotes}

\icmltitlerunning{Deep Learning for EPL Performance Forecasting}

\begin{document}

\twocolumn[
\icmltitle{Deep Learning and Transfer Learning Architectures \\ for English Premier League Player Performance Forecasting}




\begin{icmlauthorlist}
\icmlauthor{Daniel Frees*}{yyy}
\icmlauthor{Pranav Ravella}{yyy}
\icmlauthor{Charlie Zhang}{yyy}

\end{icmlauthorlist}

\icmlaffiliation{yyy}{Stanford University}

\icmlcorrespondingauthor{Daniel Frees}{dfrees@stanford.edu}
\icmlcorrespondingauthor{Pranav Ravella}{pravella@stanford.edu}
\icmlcorrespondingauthor{Charlie Zhang}{czzq@stanford.edu}

\icmlkeywords{Machine Learning, Convolutional Neural Networks, Soccer, Forecasting, Time Series, Transfer Learning, Boosting, Transformers, Natural Language Processing}

\vskip 0.3in
]



\printAffiliationsAndNotice{\icmlEqualContribution} 

\begin{abstract}
This paper presents a groundbreaking model for forecasting English Premier League (EPL) player performance using convolutional neural networks (CNNs). We evaluate Ridge regression, LightGBM and CNNs on the task of predicting upcoming player FPL score based on historical FPL data over the previous weeks. Our baseline models, Ridge regression and LightGBM, achieve solid performance and emphasize the importance of recent FPL points, influence, creativity, threat, and playtime in predicting EPL player performances. Our optimal CNN architecture achieves better performance with fewer input features and even outperforms the best previous EPL player performance forecasting models in the literature. The optimal CNN architecture also achieves very strong Spearman correlation with player rankings, indicating its strong implications for supporting the development of FPL artificial intelligence (AI) Agents and providing analysis for FPL managers. We additionally perform transfer learning experiments on soccer news data collected from The Guardian, for the same task of predicting upcoming player score, but do not identify a strong predictive signal in natural language news texts, achieving worse performance compared to both the CNN and baseline models. Overall, our CNN-based approach marks a significant advancement in EPL player performance forecasting and lays the foundation for transfer learning to other EPL prediction tasks such as win-loss odds for sports betting and the development of cutting-edge FPL AI Agents.
\end{abstract}

\section{Introduction} 
\label{submission}

Soccer is the most popular sport in the world, and most soccer fans agree that the greatest of its many leagues is the English Premier League (EPL). The Premier League is so popular that over 11 million people sign up each year to manage their own ‘Fantasy’ Premier League (FPL) team, selecting players each week based on their expected performance \cite{fantasyfootballreports}. FPL managers then score points according to the real-world performance of their selected players that week. For example, if a midfielder such as James Maddison gets an assist and a goal, he scores $3 + 5$ = $8$ points plus other points for clean sheets and other positive contributions. It is natural to ask whether machine learning can competitively model FPL player performance measured through the proxy of FPL points.

The task of predicting a single players' upcoming performance is challenging, especially because soccer is a low-scoring sport with very high variation in possible outcomes \cite{bunker2022application}. However, achieving solid performance on this fine-grained regression task could translate into much better performance on the simpler tasks of player selection (ranking) and match outcome prediction ($3$-class classification).  As a result, developing a superior player forecasting model has implications for FPL artificial intelligence (AI) agents, profiting from sports betting by winning against official betting agencies in expectation, and more. 

Here we construct several model architectures with the goal of predicting upcoming player match performance. In predicting a player $p$'s upcoming gameweek performance, our input data consists of tabular player performance data from the prior gameweeks. For our transfer learning models, we also input recent news corpus data mentioning $p$ leading up to the upcoming gameweek. Since FPL points are scored differently for each position, four models are trained for each architecture: goalkeeper (GK), defender (DEF), midfielder (MID), and forward (FWD) models. 

\section{Related Works}\label{relworks}

\subsection{Soccer Match Outcome Binary Classification}

Researchers have applied machine learning to soccer forecasting problems extensively over the last decade. EPL match prediction was a very popular research area in the mid 2010s, with researchers investigating Poisson processes, Bayesian networks, graph modeling, and other techniques with success \cite{koopman2013modeling, razali2017premier, grund2012premier, baboota2018various}. For example, \cite{koopman2013modeling} was able to successfully model EPL match results with Poisson processes and beat betting odds at sports agencies to win money. While these models are strong, they model a simpler problem than player performance forecasting. 

\subsection{Cutting Edge Player Performance Forecasting}

More recently, modern machine learning approaches stemming from the field of natural language processing (NLP) such as LSTMs, designed for learning sequential data, have been proposed as strong candidates for modeling EPL player performance since they have the capability to model complex patterns in time-series data \cite{lindberg2020lstm}. While LSTMs are strong model candidates for time-series data, the most recent research suggests that one-dimensional (1D) CNNs may be more powerful due to their increased flexibility in modeling time patterns through filters \cite{wibawa2022timeseries}. In fact, CNNs have been shown to be a feasible architecture for EPL player performance forecasting, though the original research in this area did not experiment widely to find an optimal CNN architecture, and failed to report test errors \cite{ramdas2022cnn}.

\subsection{Natural Language Signals}

The idea that predictive signals for upcoming soccer match results could be obtained via text data is not new; in fact several researchers have investigated the usage of Twitter text data for this task \cite{godin2019using, schumaker2016twitter}. Godin had great success, managing a 30\% profit by betting laterally to official bookmakers using his Twitter aggregation model. However, no literature to our knowledge has investigated the use of a news corpus for soccer prediction, nor has the efficacy of text data been evaluated for the more challenging task of player performance forecasting.

\section{Datasets and Features}

We scraped $2020-2021$ and $2021-2022$ season EPL data from \texttt{vaastav/Fantasy-Premier-League} \cite{fpl_repo}. Fuzzy matching was used to clean up variations found in player names so that data could be matched together. Slight mismatches were typically the result of abbreviation or variations in diacritics. Each player's data was cleaned using \texttt{pandas} \cite{pandas} and augmented with a column for upcoming match difficulties, feature engineered from official difficulty ratings for EPL teams each season. Benched players with no minutes played were dropped. Notably, if benched players are \textit{not} dropped, model performance is skewed high since predicting these players to score $0$ points becomes nearly trivial for all models investigated here to learn. After dropping benched players, data was organized by position (GK, DEF, MID, FWD). Time-series data was discretized into windows of size $w$. For the CNN, the entire $w \times f$ window of $f$ features is retained to allow for pattern learning in the CNN filters. We denote this feature-window $X^{(i)}$. For baseline modeling, a sliding average of $w$ weeks is used to generate a sliding average feature vector $x^{(i)}$. The target data $y^{(i)}$ and upcoming match difficulty $d^{(i)}$ are collected from the week following the current window. 

\begin{figure}[ht]
    \vskip 0.1in
    \begin{center}
        \includegraphics[width=1\columnwidth]{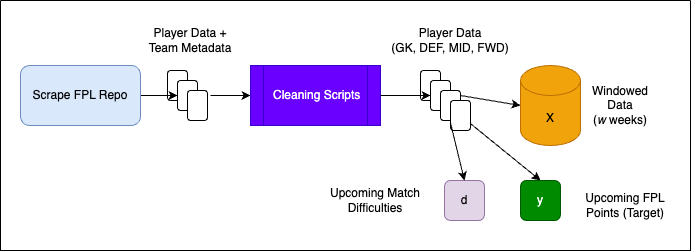}
        \caption{High-Level Data Scraping and Cleaning Pipeline.}
        \label{eda_pos_dist}
    \end{center}
\end{figure}

Based on stability of validation errors in initial experiments, we settled on split sizes of approximately $60$\% training, $25$\% validation, and $15$\% testing data. Splits were performed player-by-player to avoid data leakage.

\begin{table}[ht]
  \centering
  \caption{Dataset Size by Model (\# of Examples) \\ \tiny Note: Transfer model trained on fewer data due to compute limitations \\}
  \label{tab:your_label}
    
  \begin{tabular}{p{2cm}cccc}
    \toprule
    \cmidrule(lr){1-3}
     & Train & Validation & Test   \\
    \midrule
    GK & 1272 & 558 & 429 \\
    DEF &  6016 & 2548 & 1479 \\
    MID  & 6753 & 2993 & 1839 \\
    FWD  & 2437 & 963 & 735 \\
    \bottomrule
  \end{tabular}
\end{table}

Notably, some players are much more challenging to forecast than others (eg. a superstar like Erling Haaland will fluctuate in score more wildly than a solid CDM like Yves Bissouma). As a result, some splits end up with more predictable players and validation error and test error can wind up lower than train error. To mitigate this, we stratified on player skill (discretized from an engineered helper feature \texttt{avg\_score}). Cross-validation (CV) demonstrated improved average performance with skill stratification, and more consistency between splits (\texttt{stdev\_score} stratification performed equivalently, likely due to the heterscedasticity of player points residuals across \texttt{avg\_score}). 

Standard scaling ($z = \frac{x - \mu}{\sigma}$) was applied to the CNN and Ridge Regression input data, but not to the LightGBM. LightGBM was the primary comparison model competing with our custom CNN, and tree-based models are invariant to scaling. 

\begin{table}[!ht]
  \centering
  \caption{Example $X^{(i)}$ with $w=2$}
  \label{tab:player_stats}
    
  \begin{tabular}{llll}
    \toprule
    goals\_scored & assists & total\_points & name \\
    \midrule
    0 & 0 & 1 & Aleks. Mitrović \\
    2 & 0 & 12 & Aleks. Mitrović \\
    \bottomrule
  \end{tabular}
\end{table}

\begin{table}[!ht]
  \centering
  \caption{Example $d^{(i)}, y^{(i)}$ corresponding to the $X^{(i)}$ in \autoref{tab:player_stats}}
  \label{tab:player_stats}
    
  \begin{tabular}{cc}
    \toprule
    Upcoming Match Difficulty ($d^{(i)})$ & Target Score ($y^{(i)}$) \\
    \midrule
    -1 & 2 \\
    \bottomrule
  \end{tabular}
\end{table}

\subsection{Transfer Learning Dataset}

\begin{figure}[ht]
    \begin{center}
        \includegraphics[width=0.65\columnwidth]{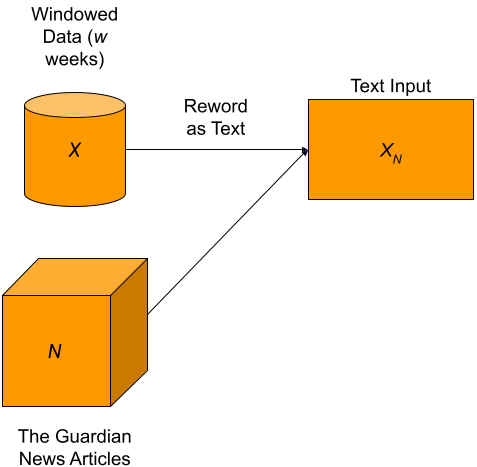}
        \caption{Expansion of Data Pipeline for Transfer Learning per Position.}
        \label{data_flow_Tl}
    \end{center}
\end{figure}

Additional steps were required to construct the transfer learning dataset. New corpus data for each player was scraped from The Guardian \cite{guardian}. An average of $\approx950$ articles were found per player, but some players were discussed much more than others ($\text{stdev} \approx 900)$. For each example, the three most recent articles referencing the player prior to the upcoming game were collected. Tabular windowed data ($X$) was reworded as a sentence and appended into each text example alongisde the first 512 words of each article ($N$) to form each example $X_N$ (\autoref{data_flow_Tl} ).

\subsection{Data EDA}

\section{Methods}

\subsection{Baseline Models}


For our first baseline model, we chose a simple regression algorithm: Ridge regression. For a more complex baseline, we evaluated gradient boosting. Gradient boosting builds a sequence of trees in which the next tree in the sequence is fit to the residuals of the current ensemble. In other words, let $T(x; \Theta_m)$ be the $m$th tree in the ensemble, then its prediction target should be the negative gradient, $-g_m$, at iteration $m$, where 

\begin{equation*}
\Theta_m = \arg \min_{\Theta} \sum_i^N (-g_{im} - T(x_i; \Theta))^2 
\end{equation*}

In essence, it is a ensemble method whose output $f(x)$ can be expressed as a combination of the predictions of multiple weak learners $f_M(x) = \eta \sum_m^M T(x; \Theta_m)$. The implementation of gradient boosting we use here is LightGBM \cite{ke2017lightgbm}. Compared to other implementations, LightGBM grows trees leaf-wise (as opposed to depth-wise), splitting the leaf node with the highest gain across all depth levels \cite{bestfirst}. 


\subsection{Deep Learning Models}

\begin{figure}[!h]
    \begin{center}
        \includegraphics[width=0.95\columnwidth]{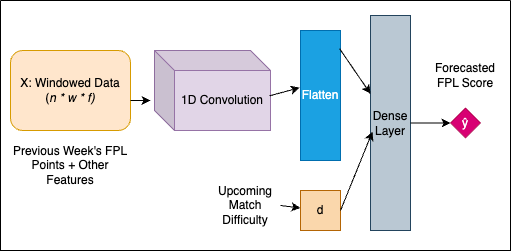}
        \caption{Custom CNN Architecture for FPL Performance Forecasting.}
        \label{cnn_arch}
    \end{center}
\end{figure}

For our deep learning model, we decided to architect a custom one-dimensional convolutional neural network (1D CNN) using Tensorflow \cite{tensorflow2015} to enable learning of complex time-series patterns by using backpropagation to derive filter weights. The model was constructed iteratively through eleven versions, performing grid searches for optimal hyperparameters and evaluating averaged top-$10$ lowest validation MSEs at each iteration to choose the best architecture going forward. Our optimal configuration consisted of a single one-dimensional convolution layer, which is flattened and concatenated with upcoming match difficulty before being passed through a single dense hidden layer, and then to the final layer (\autoref{cnn_arch}).

Our cost function for this CNN consisted of MSE with ElasticNet (L1L2) regularization on both the convolution weight matrices ($C$) and the dense layer weight matrix ($W^{[1]}$), with L1 strength determined by $\lambda_1$ and L2 strength determined by $\lambda_2$. Thus, our cost function for batch size $B$ was as follows: 
\begin{align*}
J_{CNN} & = \left(\frac{1}{B}\sum_{k=1}^B (y^{(i)} - \hat{y^{(i)}}) ^2 \right) \\
& + \lambda_1 \left(||C||_1 + ||W^{[1]}||_1 \right) \\
 & + \lambda_2 \left(||C||_2^2 + ||W^{[1]}||_2^2 \right)
\end{align*}

To train our CNN, we used backpropagation to calculate gradients with respect to our cost function, and the Adam \cite{kingma2014adam} optimizer for weight updates. Early stopping was employed to further counteract overfitting.

\subsection{Transfer Learning Model}

Forecasting player performance using news corpus data boils down to a text regression problem. Scoring text using transformers has proven to be one of the best methods for tasks such as these \cite{transformersNLP}. A transformer network uses attention blocks to model token connections in input sequences (in the form of directional soft weights between tokens) and output embeddings. These embeddings can then be input into a fully-connected layer to output logits (real numbers in the embedding space). For classification, a softmax activation is used to convert the logits into probabilities to predict the next best word \cite{transformersNLP}, whereas for regression we project the output to a single real-valued score. The self-attention mechanism core to transformers enables them to understand context and perform extremely well in understanding signals from longform texts such as news articles. Given the length of our news corpus data, a longformer model from AllenAI with a sequence length of 4096 was chosen for the transfer learning regression. Most transformer-based models are unable to process long sequences due to the attention mechanism that scales quadratically. Longformer uses a sliding-window attention mechanism to do this in linear time \cite{beltagy2020longformer}.

Since we desire a real number output, we want a single output logit from the regression. Our selected HuggingFace model \cite{wolf2020huggingfaces} applies sigmoid activation to output logits, so we appended a simple scaling layer to project the sigmoid output to a reasonable range of possible FPL scores (between $-5$ and $24$). This architecture is not ideal because it attempts to train a scaled classification problem as regression. However, compute constraints meant that we could not freeze layers up to the logits and retrain a fully-connected layer. Our goal here was merely to identify whether there is any learnable signal from these longform texts. Should there prove to be a predictive signal, it would make sense to purchase a GPU cluster to learn weights more deeply back into the longformer network and achieve better performance.

Each input example $X_N$ was tokenized into $4096$ tokens for input into our longformer model. Each model (GK, DEF, MID, FWD) took around four to seven hours to train for 25 epochs using an NVIDIA A100 with PyTorch \cite{NEURIPS2019_9015}, so tuning was limited and data was restricted to the $2021-22$ season.

\section{Experiments and Results}
We used mean squared error (MSE) as our cost function to evaluate our baseline, CNN, and transfer learning models.

\begin{table}[!ht]
  \centering
  \caption{Holdout (Test) MSE for Optimal Architectures}
  \label{tab:optimal}
    
  \begin{tabular}{p{2cm}cccc}
    \toprule
    \multirow{2}{*}{Model} & \multicolumn{4}{c}{Architecture} \\
    \cmidrule(lr){2-5}
     & Ridge & LGBM & CNN & Transfer*  \\
    \midrule
    GK & 6.46 & 6.22 & 5.08  & 8.22\\
    DEF &  7.20 & 7.24 & 5.87 & 9.66\\
    MID  & 6.08 & 6.11 & 6.16 & 8.40\\
    FWD  & 7.19 & 7.28 & 6.22 & 10.12\\
    \textbf{AVG} & \textbf{6.73} & \textbf{6.71} & \textbf{5.83} & \textbf{9.10}\\
    \bottomrule
  \end{tabular}
\tiny *Trained only on 2021-2022 due to GPU limitations
\end{table}

\subsection{Baseline ML Experiments}
The experiments for Ridge regression and gradient boosting mostly consisted of hyperparameter tuning. We used grid search with stratified 5-fold CV \cite{scikit-learn}. This is particularly critical for gradient boosting as it has numerous hyperparameters to inhibit excessive model complexity. The hyperparameter search space can grow exponentially, so we iteratively performed several rounds of grid search to rule out unreasonable options.

\begin{figure}[!ht]
\begin{center}
\centerline{\includegraphics[width=0.9\columnwidth]{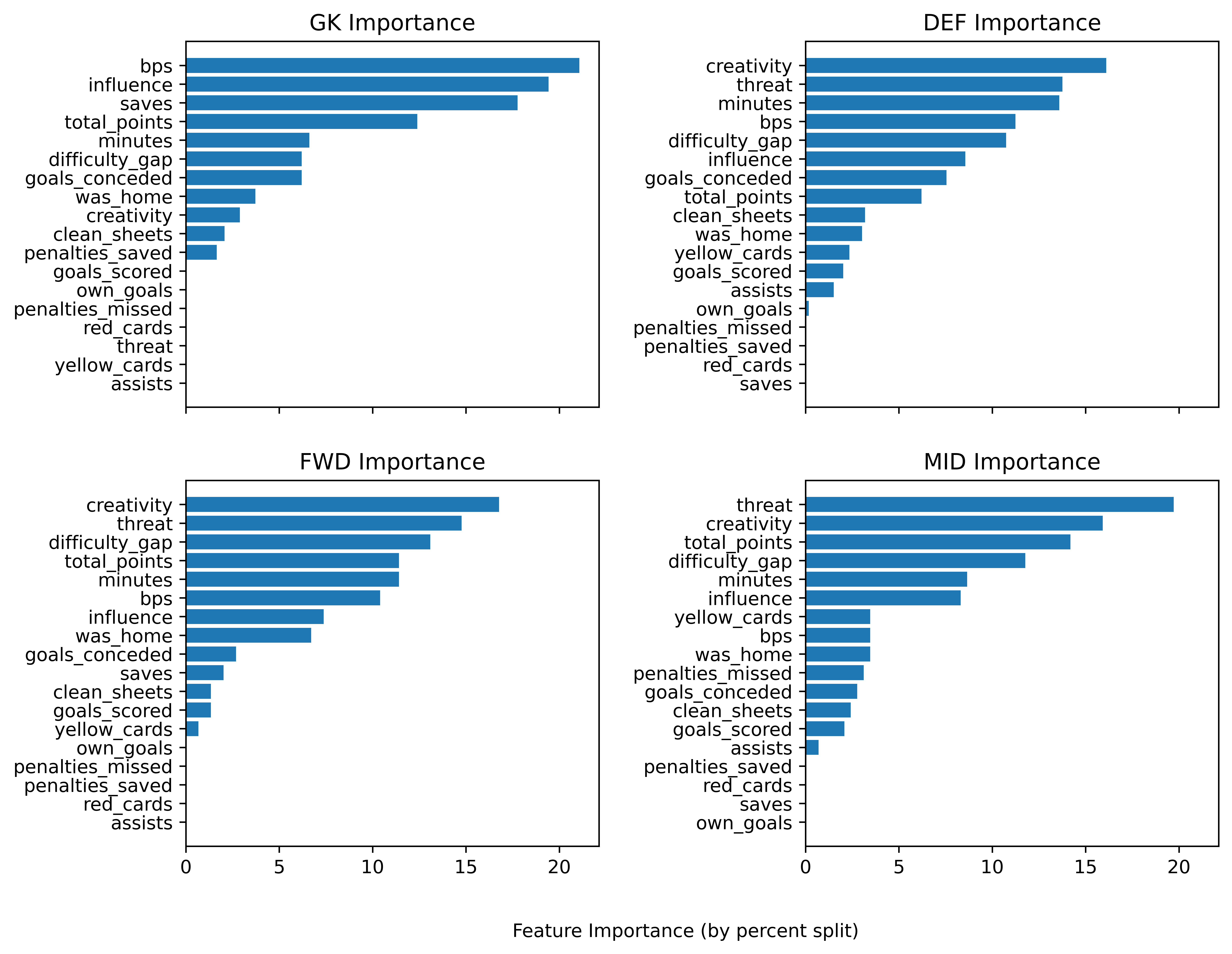}}
\caption{LightGBM feature importance measured by the percentage of splits performed with a given feature.}
\label{lgbm_imp}
\end{center}
\end{figure}

\begin{figure}[!ht]
\begin{center}
\centerline{\includegraphics[width=0.9\columnwidth]{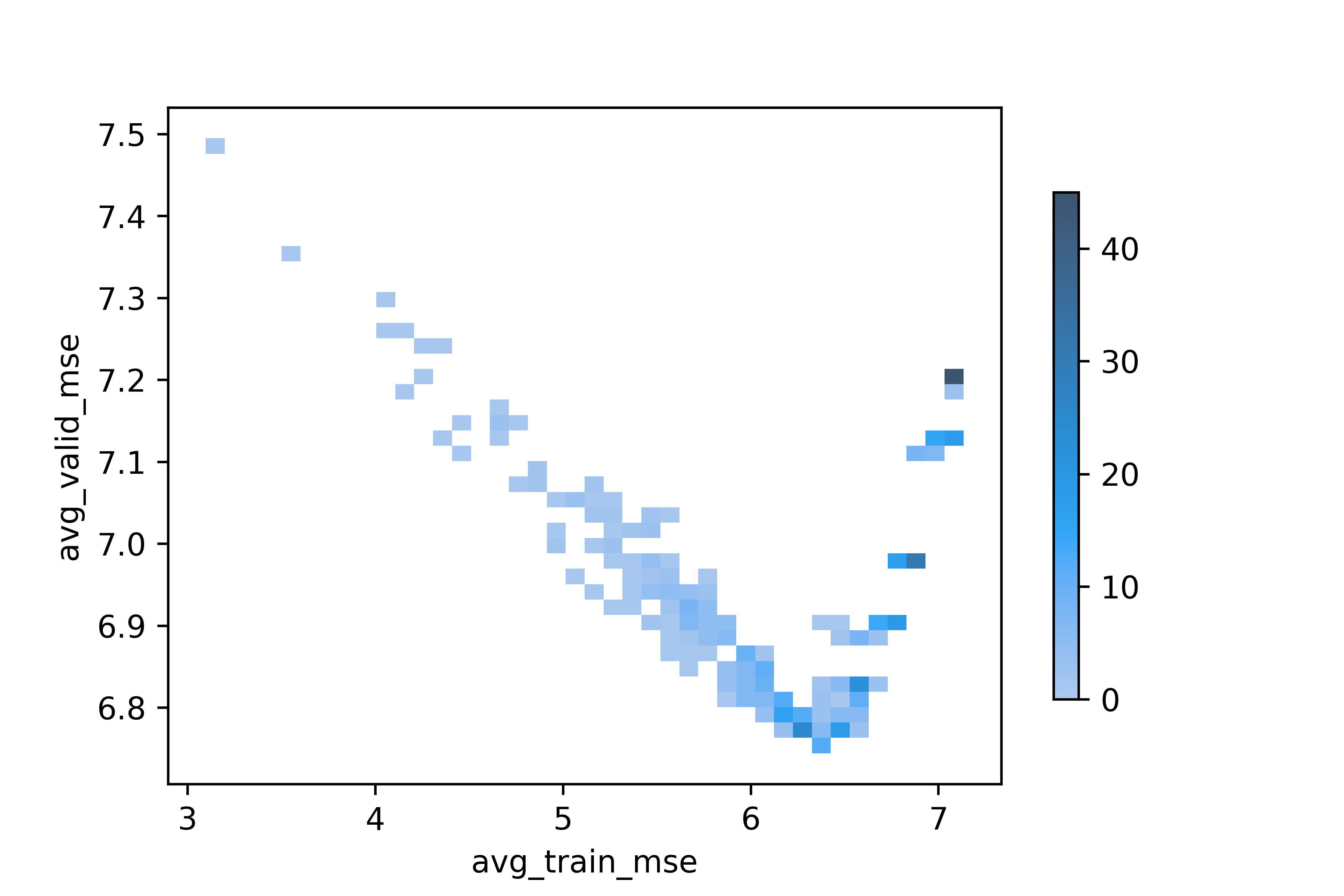}}
\caption{2D Histogram on the average performance of LightGBM during CV for each combination in the hyperparameter grid.}
\label{overfit_lgbm}
\end{center}
\end{figure}

In Figure \ref{overfit_lgbm}, we observed that even with the adjusted search space, multiple combinations were still vulnerable to overfitting of varying severity. How does the best set of hyperparameters achieve the optimal bias-variance tradeoff? On the ensemble level, only 50 trees are fit. For each tree, we had only a depth of 3, an L2 regularization strength of 10, and 7 leaf nodes each holding 70 observations minimum. As our dataset was smaller than a typical application scenario that calls for LightGBM, CV ended up favoring hyperparameters that more aggressively limit tree complexity.

\begin{figure}[!ht]
\begin{center}
\centerline{\includegraphics[width=0.95\columnwidth]{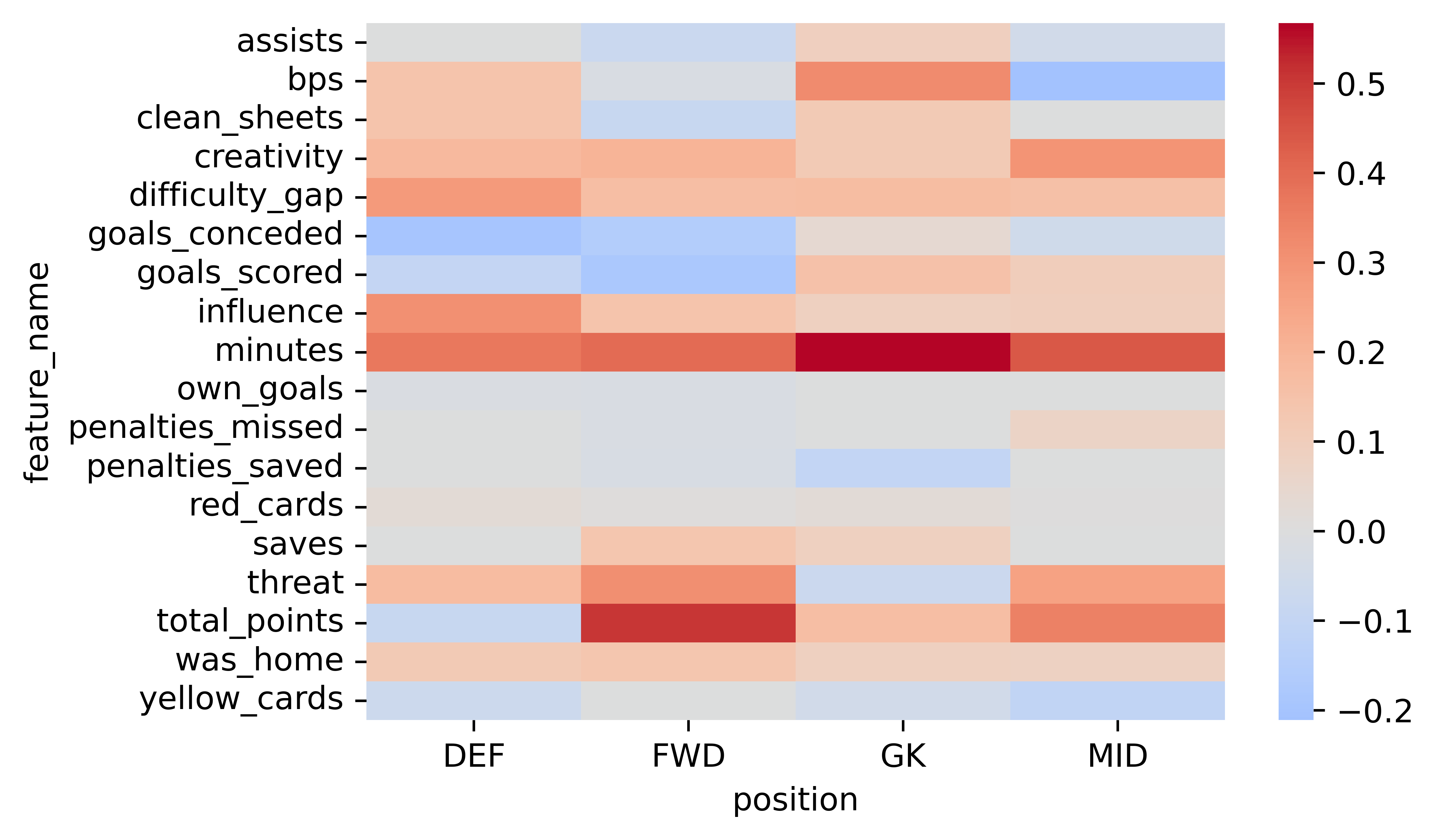}}
\caption{Heatmap of Ridge Regression coefficients. Models have the same regularization strength. Warm colors represent positive coefficients and vice versa.}
\label{lr_imp}
\end{center}
\end{figure}

\begin{figure}[!ht]
\begin{center}
\centerline{\includegraphics[width=0.95\columnwidth]{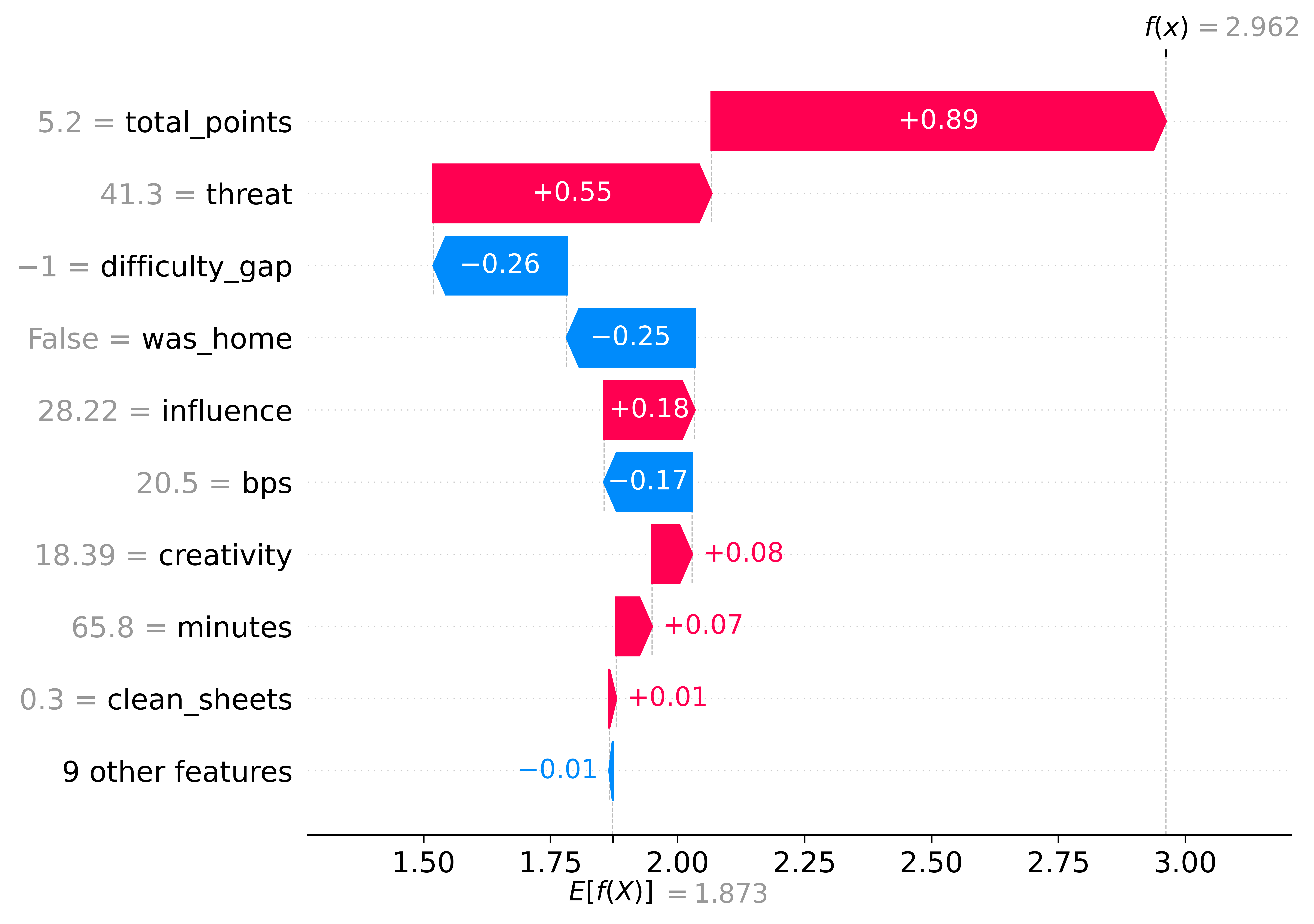}}
\caption{LightGBM feature contribution visualized with Shapley values. The starting point $E[f(x)]$ represents the average (FWD) model output with the testing set. The ending point $f(x)$ represents the model output for a given observation with feature values specified on the left side.}
\label{lgbm_shap}
\end{center}
\end{figure}

There are multiple ways we can interpret baseline model outputs and feature importance. For Ridge Regression, we can compare the coefficients, shown as a heatmap in Figure \ref{lr_imp}. For LightGBM, we can examine in Figure \ref{lgbm_imp} what percentage of all the splits is based on a given feature as a proxy for its importance. Alternatively, visualized in Figure \ref{lgbm_shap}, Shapley values can illustrate each feature’s numerical contribution to the final tree ensemble output, adding some much needed explainability \cite{shapexplain}. First, we observe several features that are consistently important for both baseline models. Those include, for example, \textit{minutes} (as players with more playtime have more opportunities to score points and are more likely recognized by the team to be on the starting lineup), \textit{difficulty\_gap} (as the opponent’s relative strength to the player’s team can translate to ball possession percentages and pressure), \textit{influence/creativity/threat} (as these FPL-compiled metrics take into account finer-grained match-events beyond low occurrence events such as goals and assists), and \textit{total\_points} (as this sliding-window average of recent weeks’ points is a fair indicator of a player’s recent form). Subsequently, we notice the effect of modeling each position separately. For example, the LightGBM goalkeeper model values \textit{saves} a lot and, compared to other positions, relies much less on \textit{creativity/threat} (which focus on actions building up to goals) compared to \textit{influence}. \\

A number of the regression coefficients agree in directionality of effect across positions. This breaks down quote strongly for \textit{bps} (bonus points), \textit{goals\_scored}, \textit{goals\_conceded}, \textit{clean\_sheets}, \textit{threat} and \textit{total\_points}. The predictor for which the coefficient trend varies most across positions is \textit{goals\_scored} (sliding window average goals in recent weeks) for forwards. One possible explanation is that scoring goals is such a rare event for individual players that, as a result, scoring goals is often followed by a week without goals and with low points. Our results do seem to suggest that goal-scoring is not an independent event week-to-week. Whether the negative effect of previous week goals on current week goals for forwards is an element of psychology, team management, or something else entirely is not within the scope of this analysis, but it does call for further research. 

From Table \ref{tab:optimal} and \ref{baseline_res} it is noticeable that gradient boosting consistently achieved lower MSE on the training set yet tied against Ridge regression during validation and testing. This could be an indicator for slight overfitting despite the grid search to balance fit and generalizability. As a result, Ridge regression is favorable for its superior interpretability. However, in the future, if more advanced features with interplay are introduced, (such as tactics, team formations, and player movement on the field), gradient boosting may have a more definitive performance advantage. 

\begin{table}[!ht]
\caption{Avg MSE during CV for baseline models}
\label{baseline_res}
\begin{center}
\begin{small}
\begin{sc}
\begin{tabular}{lcccr}
\toprule
\multirow{2}{*}{Model} & \multicolumn{2}{c}{Ridge} & \multicolumn{2}{c}{LightGBM}\\
    \cmidrule(lr){2-5}
     & Train & Valid & Train & Valid \\
\midrule
GK     & 6.59 & \textbf{6.65} & 6.34 & 6.68 \\
DEF    & 7.11 & 7.14 & 6.69 & \textbf{7.12} \\
MID    & 6.12 & \textbf{6.14} & 5.87 & 6.18 \\
FWD    & 6.75 & 6.77 & 6.41 & \textbf{6.73} \\
\bottomrule
\end{tabular}
\end{sc}
\end{small}
\end{center}
\end{table}

\subsection{Deep Learning Experiments}

\subsubsection{Hardware}
All CNN training was performed using TensorFlow \cite{tensorflow2015} on a $14$ core Apple Silicon M3 Max CPU.

\subsubsection{Hyperparameters}
Experiments were performed using learning curves to determine optimal training hyperparameters of \textbf{250} epochs, learning rate of \textbf{0.001}, batch size of \textbf{32} and early stopping tolerance and patience of $\bm{1 \times 10^{-4}}$ and \textbf{20} iterations, respectively. A custom implementation of grid search was developed to determine optimal window size, kernel size, number of filters, number of dense neurons, activation functions, amount of numerical features to include, and stratification strategy for GK, DEF, MID, and FWD models. Subsequent experiments reduced the hyperparameter search space by fixing certain optimal hyperparameters based on the results of previous experiment iterations.

\subsubsection{Architecture}
\begin{figure}[]

    \centering
        \includegraphics[width=0.95\columnwidth]{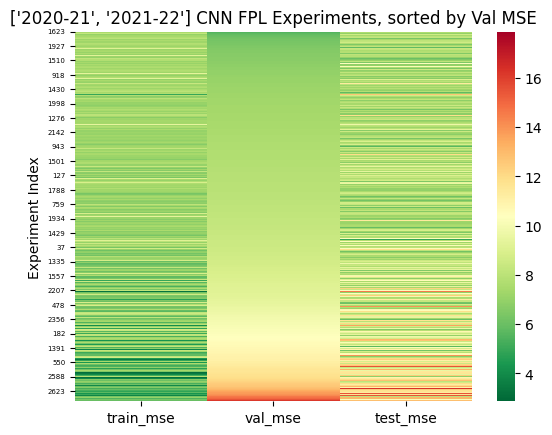}
        \caption{Large grid search for CNN v6. Train, Val, Test MSEs.}
        \label{fig:v6_gridsearch}
\end{figure}

Model architecture was refined iteratively through the first $7$ CNN grid search experiments, yielding the architecture described in \textbf{4.2}. This architecture was selected based on decreased average validation MSE, as well as reduced upper bound on validation MSE. For example, between our v6 architecture experiments $\left( \text{see \autoref{fig:v6_gridsearch}} \right)$, and v11 architecture experiments, the upper bound on validation MSE fell approximately $50$\%, and average validation MSE fell $20$\%. 

\subsubsection{Optimal Architecture and Parameters}

The optimal CNN models, selected by lowest validation MSE, achieved a holdout MSE of \textbf{5.08} for goalkeepers, \textbf{5.87} for defenders, \textbf{6.16} for midfielders, and \textbf{6.22} for forwards. The top 5 averaged CNN models achieved a holdout MSE of \textbf{6.08} for goalkeepers, \textbf{6.50} for defenders, \textbf{6.27} for midfielders, and \textbf{6.59} for forwards. An abridged view of their hyperparameter configurations can be found in \autoref{tab:cnn_config}.

\begin{table}[!ht]
\hspace*{-6mm}
    \caption{Optimal configuration for each CNN model (abridged)}
    \label{tab:cnn_config}
    \begin{center}
        \begin{small}
            \begin{sc}
                \begin{tabular}{lccc}
                    \toprule
                    \multirow{2}{*}{Position} & \multicolumn{2}{c}{\# Weeks} & \multirow{2}{*}{Amt\_Num\_Features} \\
                    \cmidrule(lr){2-3}
                    & Window & Kernel \\
                    \midrule
                    GK  & 6 & 1 & ptsonly \\
                    DEF & 9 & 1 & ptsonly \\
                    MID & 3 & 2 & ptsonly \\
                    FWD & 9 & 1 & ptsonly \\
                    \bottomrule
                \end{tabular}
            \end{sc}
        \end{small}
    \end{center}
\end{table}

\subsubsection{Feature Importances}

As seen in \autoref{tab:cnn_config}, only the previous $w$ weeks of FPL points and upcoming match difficulty $d$ were critical for the CNN to best predict a player's points in the upcoming week. 

Notably, including \texttt{ict\_index} (containing information about influence, creativity, and threat) did not benefit the CNN model, despite the importance of each of these features in the optimal LightGBM models. \texttt{minutes} were important in the optimal trained Ridge regression models (as selected by cross validation), but also did not prove to be an important feature for the CNN.

\subsubsection{Qualitative Analysis}\label{qual}

Qualitatively, we observe reasonable results. Taking a closer look at the best and worst examples for the optimized CNN MID model, we see that the two worst predictions (\autoref{bestandworst}) are made on outstanding outlier performances of $21$ points, likely via hat tricks plus bonus points. The best two examples (\autoref{bestandworst}) occur where the CNN predicts relatively poor performance based on low-scoring previous weeks, and the player performs as predicted. The CNN model makes reasonable predictions, but is unsurprisingly unable to forecast outliers.

\begin{table}[!ht]
    \centering
    \caption{Best and Worst $2$ examples by MSE (CNN MID model).}
    \label{bestandworst}
    \begin{tabular}{ccccccc}
        \toprule
        \multicolumn{2}{c}{\textbf{Points}} & \multirow{2}{*}{MSE} & \multirow{2}{*}{$d^{(i)}$} & \multicolumn{3}{c}{Points Previous Weeks} \\
        \cmidrule(lr){1-2}\cmidrule(lr){5-7}
        True & Pred & & & Wk 0 & Wk 1 & Wk 2 \\
        \midrule
        21 & 2.2 & 352.8 & -2.0 & 2 & 3  & 2\\
        21 & 4.6 & 270.4 & 2.0 & 3 & 15 & 12\\
        \cmidrule(r){1-7}
        2 & 2.0 & $\approx 0$ & 0.0 & 2 & 6 & 1 \\
        1 & 1.0 & $\approx 0$ & 3.0 & 0 & 1 & 0 \\
        \bottomrule
    \end{tabular}
\end{table}

\subsubsection{Predictions vs. True Performances}

While our CNN models achieve great performance relative to Ridge regression, LightGBM, and previous deep learning models in the literature \cite{lindberg2020lstm}, they fail to accurately predict outliers. As seen in \autoref{bestandworst}, the worst errors for the midfielder model occurred when players had standout performances. Below, we visualize this general failure to predict outliers across each CNN model. Most likely, training punishes outlier model predictions due to their rarity. As seen in the plots, player gameweeks with points between $0$ and $3$ constitute the majority of the data. As such, unless the CNN is extremely certain a player is going to have an outlier performance, it is better to predict within the typical range of player performance values. On average, a wrong outlier prediction will add a greater error since on average, player performances fall within the aforementioned low-score range. 

\begin{figure}[!htb]
    \begin{center}
        \caption{Goalkeeper Predictions vs. True Performance (Holdout Data)}
        \includegraphics[width=0.9\columnwidth]{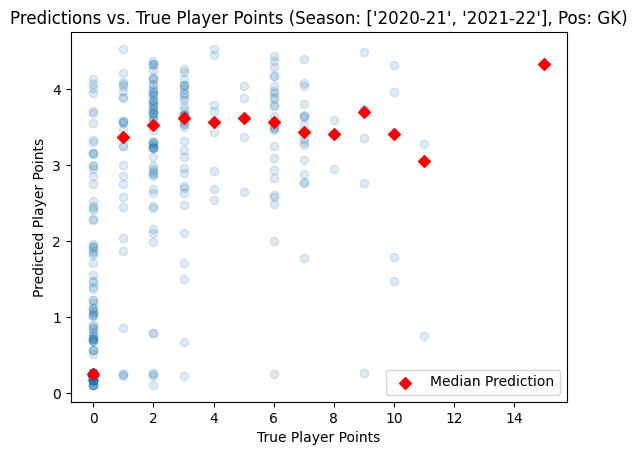}
        \label{fig:gk_pred_true}
    \end{center}
\end{figure}

\begin{figure}[!htb]
    \begin{center}
        \caption{Defender Predictions vs. True Performance (Holdout Data)}
        \includegraphics[width=0.9\columnwidth]{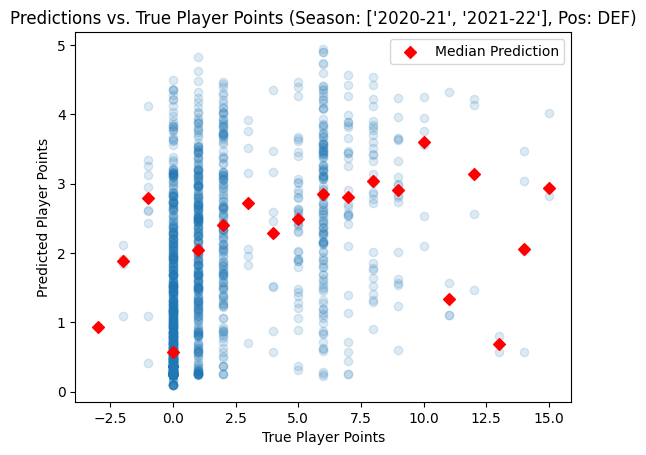}
        \label{fig:def_pred_true}
    \end{center}
\end{figure}

\begin{figure}[!htb]
    \begin{center}
        \caption{Midfielder Predictions vs. True Performance (Holdout Data)}
        \includegraphics[width=0.9\columnwidth]{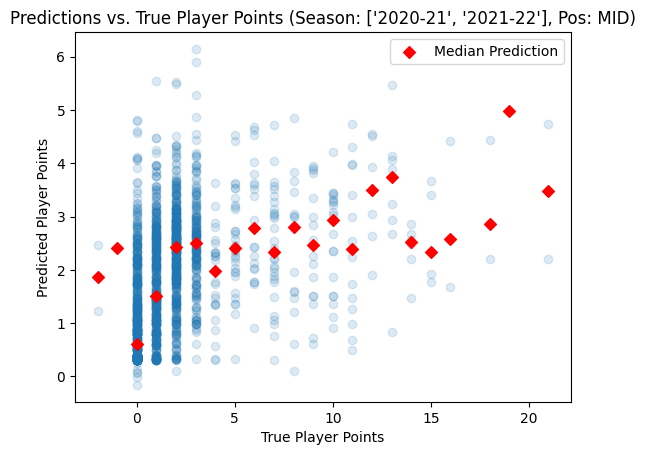}
        \label{fig:mid_pred_true}
    \end{center}
\end{figure}

\begin{figure}[!htb]
    \begin{center}
        \caption{Forward Predictions vs. True Performance (Holdout Data)}
        \includegraphics[width=0.9\columnwidth]{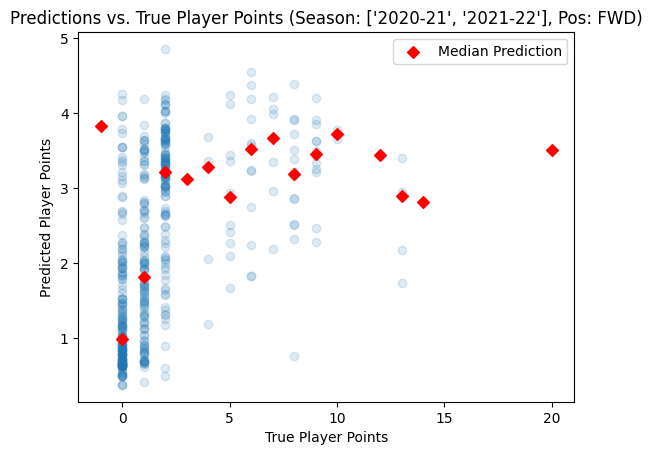}
        \label{fig:fwd_pred_true}
    \end{center}
\end{figure}

\subsubsection{Overfitting}

\begin{figure}[!htb]
    \begin{center}
        \caption{CNN v11 GK Learning Curve  \\ \tiny Adam achieves quick convergence. Validation set gets lucky split and we overfit to the validation data via early stopping. \\
}
        \includegraphics[width=0.9\columnwidth]{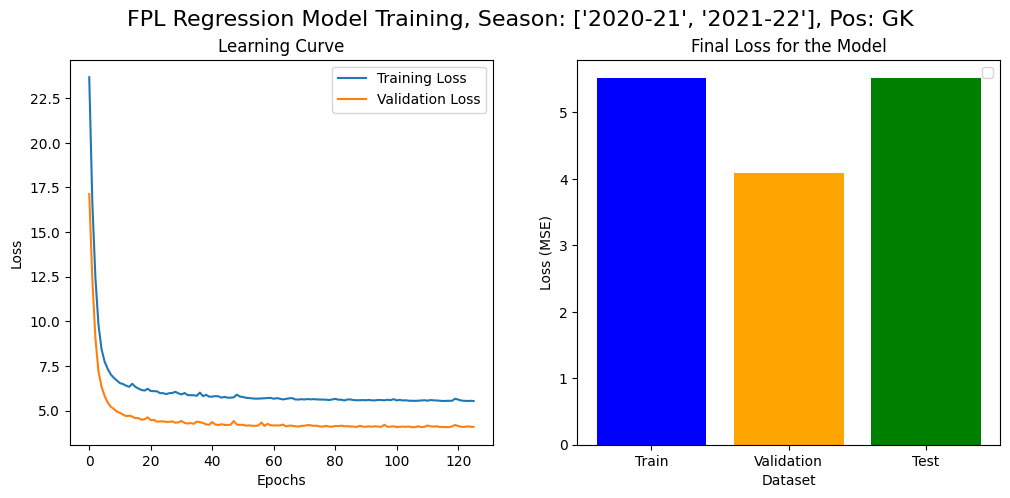}
        \label{overfit_curve}
    \end{center}
\end{figure}

Overfitting plagued early versions of the CNN, whereby validation MSE would inflect upwards as training continued. Implementing ElasticNet regularization and slower learning rate with low patience for early stopping largely mitigated this. Some overfitting lingers in the final architecture, likely the result of employing grid search with regular train/val/test shuffle splits rather than cross-validation due to compute time limitations. Notice that in \autoref{overfit_curve}, the validation data performs better than the training data (both in the first iteration prior to any learning, and in the final iteration). We hypothesize that this results from the wide variance in prediction challenge between splits, depending on the score variance of the players in the data split. Various strategies for stratified splitting were tested, including player score standard deviation (week-to-week) and player average score. Unfortunately, with a limited number of players in the EPL and the need to keep players separate across splits it was impossible to completely solve the issue. Further discussion of this anomaly can be found in the Limitations section. 

\subsection{Transfer Learning Experiments}

\subsubsection{Learning News Corpus Signal}

\begin{figure}[htb]
    \begin{center}
        \includegraphics[width=1.05\columnwidth, height=0.4\columnwidth]{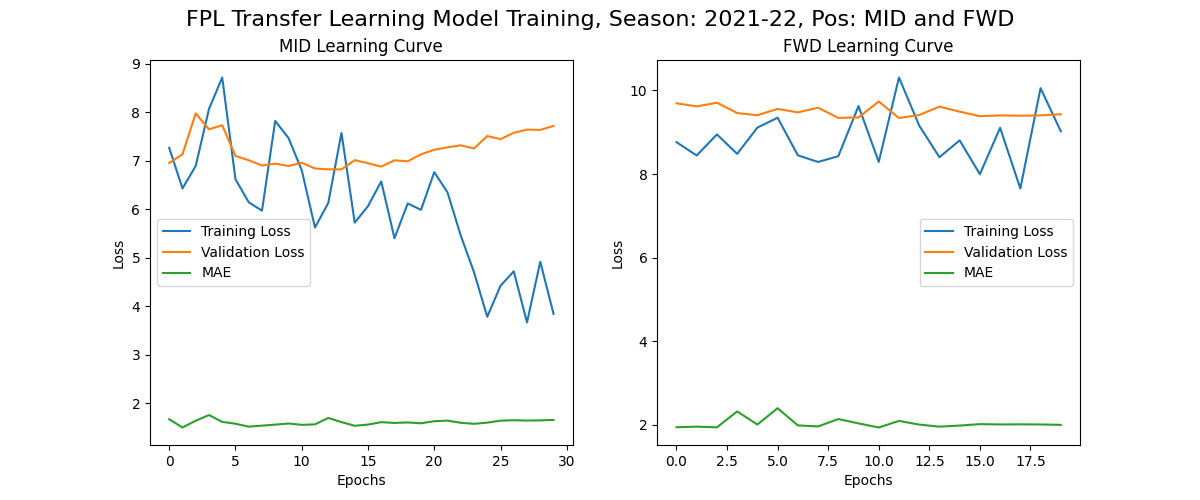}
        \caption{Learning curve and metrics for MID and FWD.}
        \label{mid_TL}
    \end{center}
\end{figure}

\autoref{mid_TL} shows the learning curves for the MID and FWD transfer learning models. GK and DEF learning curves were very similar to the MID model other than final MSE loss. As expected, the model overfit for most of the positions due to massive model parameterization compared to the relatively small dataset. As a result, the model generalizes quite poorly to the validation and test sets compared the baseline models and CNN (\autoref{tab:optimal}). The FWD model stands out in that it was unable to learn any signal in the news corpus data to improve even the training MSE (see \autoref{mid_TL}).

\subsubsection{News Corpus Data Quality}
Beyond asserting that articles occurred before upcoming kickoff time and mentioned each player, articles were not evaluated for quality in great detail. A lot of players did not have recent articles before kickoff time and as such many articles were outdated or referred to general EPL news completely unrelated to the upcoming game. 

Furthermore, predicting goal-scoring (critical for FWDs) for a particular player based on text analysis of upcoming games is harder than predicting a strong defensive performance (critical for DEFs) against a weaker team. This provides another possible justification for the inability of the transfer learning model to learn better prediction for FWDs (\autoref{mid_TL}). Overall, the news corpus did not provide a strong signal, as evidenced by the poorer performance of the transfer learning model compared with the CNN, Ridge, and LightGBM (\autoref{tab:optimal}). 

\subsection{Player Ranking Performance}

To gain a better understanding of how well our best CNN models would perform on a player ranking task, such as selecting players in Fantasy Premier League to maximize total team points, we calculate Spearman correlation between our model predictions and true player points for the holdout data. We calculate the generalized Spearman's \(\rho_s\) as follows:

{\tiny
\begin{equation*}
\hspace*{2mm}
\rho_s = \frac{S_{y\hat{y}}}{S_y S_{\hat{y}}} = \frac{
    \frac{1}{n}\sum_{i=1}^{n} \left( R(y^{(i)}) - \overline{R(y)} \right)\left( R(\hat{y}^{(i)}) - \overline{R(\hat{y})} \right)
}{
    \sqrt{
        \left( \frac{1}{n}\sum_{i=1}^{n} \left( R(y^{(i)}) - \overline{R(y)} \right)^2 \right)
        \left( \frac{1}{n}\sum_{i=1}^{n} \left( R(\hat{y}^{(i)}) - \overline{R(\hat{y})} \right)^2 \right)
    }
}
\end{equation*}
}

where:
\begin{itemize}
\item \( y^{(i)} \) is the true player points for player \( i \).
\item \( \hat{y^{(i)}} \) is the predicted player points for player \( i \).
\item \( R(y^{(i)}) \) is the rank of the true player points for player \( i \).
\item \( R(\hat{y}^{(i)}) \) is the rank of the predicted player points for player \( i \).
\item \( \overline{R(y)} \) is the mean rank of the true player points.
\item \( \overline{R(\hat{y})} \) is the mean rank of the predicted player points, averaged across all players.
\item \( S_{y\hat{y}} \) is the covariance of the true and predicted ranks.
\item \( S_y \) and \( S_{\hat{y}} \) are the standard deviations of the true ranks and predicted ranks, respectively.
\end{itemize}

Note that it is important we calculate the generalized version of Spearman's \(\rho\), which handles tied ranks, since the true player points each gameweek are always integers. This leads to many ties. 

A high Spearman correlation indicates that the predicted model points are highly monotonically related to the true player points across predicted player gameweeks. By extension, an AI agent or player using our model to select players in FPL should achieve better performance the higher the Spearman correlation. Looking at our Spearman correlation results in \autoref{tab:spearman}, we see that our optimized Ridge and LightGBM models achieve moderately high Spearman correlations and the CNN models achieve strong Spearman correlations with the true player rankings. 

Overall, our results suggest that the CNN model is also the highest performer in the player ranking task, and that the CNN model is highly promising as a foundation for developing an automatic FPL player selection tool or for providing player selection insights in a human-in-the-loop (HITL) pipeline.

\begin{table}[!htb]
\caption{Ranking Performance: Spearman's Rho for Optimal Configurations}
\label{tab:spearman}
\begin{center}
\begin{small}
\begin{sc}
\begin{tabular}{lcccc}
\toprule
 & GK & DEF & MID & FWD \\ [0.5ex] 
\midrule
Ridge     & 0.50 & 0.40 & 0.49 & 0.47 \\
LightGBM  & 0.53 & 0.40 & 0.49 & 0.48 \\
CNN          & 0.70 & 0.57 & 0.58 & 0.62 \\ [0.5ex]
\bottomrule
\end{tabular}
\end{sc}
\end{small}
\end{center}
\end{table}

\section{Conclusion}

Here we developed, to the best of our knowledge, the highest-performing model for EPL player performance forecasting in the literature. Averaged across positions, our CNN architecture outperformed our best LightGBM models by \textbf{13\%}, and the previous best model in the literature (an LSTM) by \textbf{30\%} \cite{lindberg2020lstm} \footnote{Same data source, but model evaluated on different seasons.}. Furthermore, we investigated the feasibility of forecasting performance using signals from text data collected via a news corpus, combined with transfer learning of a transformer-based model. Initial experiments did not demonstrate increased efficacy relative to our CNN forecasting model, though data and compute limitations contributed to the lower performance. Lastly, we evaluated Spearman correlation as a proxy to expected player ranking performance for our optimal Ridge, LightGBM, and CNN models, finding that the CNN models achieve highly promising ranking performance.

\subsection{Limitations}

The foremost limitation of the present work is that we were unable to completely standardize the difficulty of the performance prediction task between train/val/test splits. Because soccer is a low scoring sport, player FPL points can vary widely. For example, a talented forward player such as Son Heung-min might play very well but score no goals and receive a yellow card, netting $1$ point. The next week, Son might score a hat trick and bonus points, netting $17$ points. Predicting these high-score, low-probability events is significantly more challenging than predicting a moderately positive performance (1 assist, $5-6$ points) by a consistent playmaking midfielder such as Kevin De Bruyne. 

We attempted to standardize the prediction task across splits by stratifying our train/val/test splits by player skill (measured by average FPL points per week) and player variance (measured by standard deviatation of FPL points per week), but were unable to achieve completely similar splits. This is partly due to the fact that splits are performed player-wise to prevent data leakage, which prevents the possibility of perfect stratified splits in small player-sets, such as the EPL goalkeepers. As seen in \autoref{overfit_curve}, sometimes one split ended up with an easier prediction task. Note that splits were kept consistent across runs in comparing various hyperparameters, so that an easy validation split would not skew the hyperparameter optimization results. The variation between splits is problematic regardless as we 'leave some performance behind' in overfitting to the validation dataset, and the noisiness of split difficulty by extension means our estimates of prediction accuracy are noisy. Cross-validation combined with superior split strategy might largely mitigate this issue, and would be a great extension to this work. Cross-validation should be possible with a larger GPU cluster given the relatively small dataset.

GPU availability was also a limitation in the transfer learning experiments. Without significant external funding, we did not have access to the significant GPU compute necessary to train our longformer model sufficiently. 

\subsection{Future Directions}

Future work should focus on improving the CNN architecture proposed here via a cross-validation grid search experiment framework and more advanced stratification strategies to normalize prediction difficulty between splits. A simple next direction to start with for further stratification strategies would be to stratify splits based on average point difference week-to-week, though there may be other measures of variability that better mitigate difficulty differences between splits. 

The optimal CNN architecture presented here has strong implications for the development of FPL AI Agents and sports betting models. An AI Agent could be developed on top of our current model by leveraging budget optimization techniques alongside the CNN \cite{gupta2017ensemble}. Furthermore, sports betting models might be developed based on aggregated player predictions for each competing team. By utilizing intermediary player performance features in learning win-loss probabilities (ie. by training a Logistic Regression for wins and losses with player performances as input features), it may be possible to predict better odds than sports betting agencies, and achieve profit in expectation.

Another valuable extension to this work would be investigating the benefits of learning via direct optimization for Spearman's correlation. Recent research has developed a methodology for direct optimization of Spearman's correlation using convex hull projections \cite{blondel2020fast}. Directly optimizing for Spearman's correlation might improve player ranking performance, though this would need to be confirmed empirically, especially because ranking optimization is a novel field.

Lastly, traditional natural language understanding (NLU) techniques should be employed to directly expose sentiment, entity relationships, and other valuable NLU features attached to the relevant EPL player as features for the transformer model. Directly exposing these features  might enable more effective transfer learning by reducing noise in the input data.  

Though our minimal transfer learning example here failed to detect a signal, experimentation with a more complete transfer learning architecture might be worthwhile. To properly model the regression task, transfer learning should be conducted by freezing longformer (or other transformer model) weights, using these as a feature extractor, and learning a fully-connected layer (or a few) to predict the real-valued upcoming FPL points outcome. Transfer learning could also be improved by collecting datasets across multiple news corpuses and training with more compute power.

\section{Code}

All code for the present work can be found at: \url{https://github.com/danielfrees/mlpremier}.

\section{Acknowledgments}

  We would like to acknowledge the support of Stanford's CS229 teaching staff who helped us work through the initial iteration of this project.

\section{Funding}

  No outside funding was utilized in support of this research.

\clearpage


\bibliography{mlpremier}
\bibliographystyle{icml2023}

\newpage
\appendix
\onecolumn
\clearpage

\appendix

\section{Learned Average Filters}

Here, we briefly take a look into the (not so pretty) learned filters for the optimal CNNs to get a qualitative sense of longitudinal patterns in player performance. Note that we need to be careful about interpreting learned filters, as their effects on the final output are obstructed by the myriad of weighted connections between the flattened filters and the dense layer(s) of the CNN. 

To get a sense of excitatory (more important/ predictive) patterns in the time series data, we begin by retrieving the filter weights from the second hidden layer of the CNN model. Next, we perform z-score normalization on the filter weights to ensure they have a mean of 0 and a standard deviation of 1. Finally, we calculate the mean filter by taking the average of all (64) of the normalized filters. 

\begin{figure}[!htb]
    \vskip 0.1in
    \begin{center}
        \includegraphics[width=0.2\columnwidth]{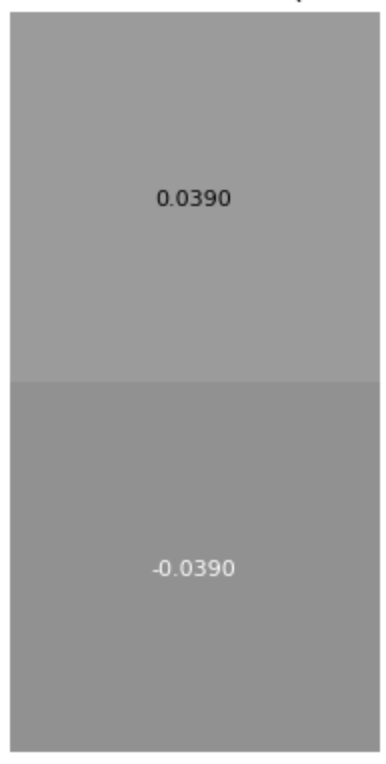}
        \caption{Averaged Normalized 1D Convolution Filter (MID Model).}
        \label{fig:cnn_mid_filter}
    \end{center}
\end{figure}

As seen in \autoref{tab:cnn_config}, the MID model was the only one which had an optimal (as selected by GridSearch) kernel with size $>1$, so we look at the average standardized filter learned for the MID model (\autoref{fig:cnn_mid_filter}). We see that there may be a very slightly more excitatory effect of the penultimate week in each application of the filter, though the average normalized filter weights are quite close to $0$. Between the confounding effects of dense layer weights and the near-zero average normalized filter weights it seems most likely that there is no consistent pattern in longitudinal feature importance. In other words, it does not seem that older weeks and more recent weeks are consistently more predictive of upcoming performance based on this initial qualitative filter analysis. 

Of course, attempting to interpret anything from such a small filter is, to begin with, a mostly futile qualitative task. 

\clearpage

\end{document}